\documentclass[letterpaper]{article} 
\usepackage{aaai24}  
\usepackage{times}  
\usepackage{helvet}  
\usepackage{courier}  
\usepackage[hyphens]{url}  
\usepackage{graphicx} 
\usepackage{natbib}  
\usepackage{caption} 
\usepackage{algorithm}
\usepackage{algorithmic}
\usepackage{booktabs}
\usepackage{amsmath}
\usepackage{amsfonts}
\usepackage{multirow}
\usepackage{xspace}

\usepackage{newfloat}
\usepackage{listings}
\newcommand{\ours}{\textbf{KGEditor}}
\DeclareCaptionStyle{ruled}{labelfont=normalfont,labelsep=colon,strut=off} 
\floatstyle{ruled}
\newfloat{listing}{tb}{lst}{}
\floatname{listing}{Listing}
\title{Editing Language Model-based Knowledge Graph Embeddings}
\author{
    Siyuan Cheng\textsuperscript{\rm 1, 2, 4}\equalcontrib,
    Ningyu Zhang\textsuperscript{\rm 1, \rm 2}\footnote{Corresponding author.}, \\
    Bozhong Tian\textsuperscript{\rm 1, 2}\equalcontrib,
    Xi Chen\textsuperscript{\rm 4}\equalcontrib, 
    Qingbin Liu\textsuperscript{\rm 4},
    Huajun Chen\textsuperscript{\rm 1, 2, 3}\footnotemark[2]
}
\affiliations{
    \textsuperscript{\rm 1}Zhejiang University\\
    \textsuperscript{\rm 2}Zhejiang University - Ant Group Joint Laboratory of Knowledge Graph\\
    \textsuperscript{\rm 3}Donghai Laboratory\\
    \textsuperscript{\rm 4}Platform and Content Group, Tencent\\

    \{sycheng, zhangningyu, tbozhong, huajunsir\}@zju.edu.cn\\
    \{jasonxchen, qingbinliu\}@tencent.com
}

\usepackage{bibentry}

\begin{document}

\maketitle

\begin{abstract}
Recently decades have witnessed the empirical success of framing Knowledge Graph (KG) embeddings via language models. However, language model-based KG embeddings are usually deployed as static artifacts, making them difficult to modify post-deployment without re-training after deployment. To address this issue, we propose a new task of editing language model-based KG embeddings in this paper. This task is designed to facilitate rapid, data-efficient updates to KG embeddings without compromising the performance of other aspects. We build four new datasets: E-FB15k237, A-FB15k237, E-WN18RR, and A-WN18RR, and evaluate several knowledge editing baselines demonstrating the limited ability of previous models to handle the proposed challenging task. We further propose a simple yet strong baseline dubbed KGEditor, which utilizes additional parametric layers of the hypernetwork to edit/add facts. Our comprehensive experimental results reveal that KGEditor excels in updating specific facts without impacting the overall performance, even when faced with limited training resources\footnote{Code and datasets are available in \url{https://github.com/zjunlp/PromptKG/tree/main/deltaKG}.}.
\end{abstract}

\section{Introduction}
\begin{figure}[htbp]
    \centering
    \includegraphics[width=0.47\textwidth]{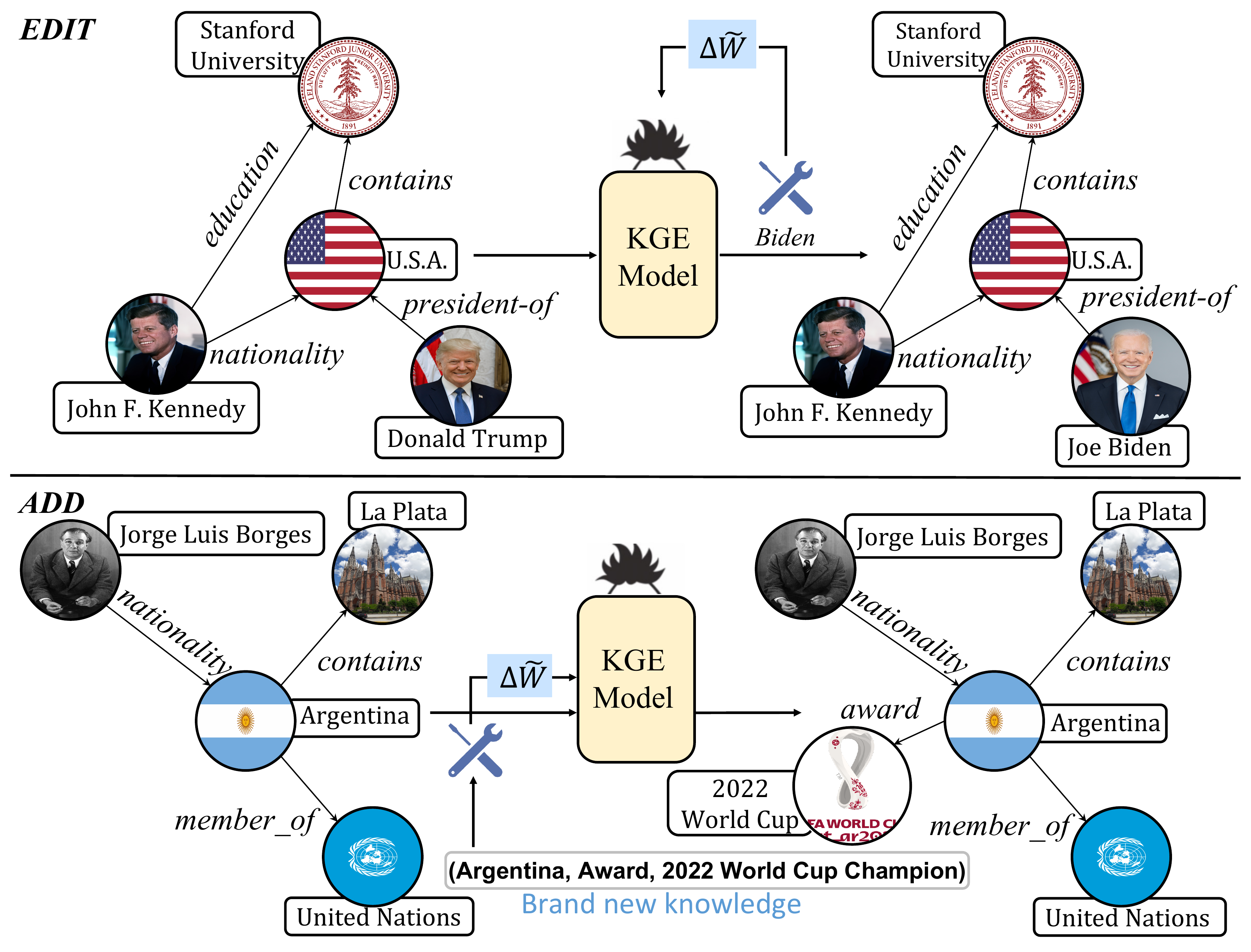}
    \caption{
    \textbf{Top}: Illustration of the EDIT task. 
    We edit the wrong fact knowledge stored in the KG embeddings. 
    \textbf{Bottom}: Illustration of the ADD task. 
    We add brand-new knowledge into the model without re-training.}
    \label{fig:intro}
\end{figure}

Knowledge Graphs (KGs) represent large-scale, multi-relational graphs containing a wealth of symbolic facts. These structures offer invaluable back-end support for various knowledge-intensive tasks, such as information retrieval, question-answering, and recommender systems. \cite{yang2021hitrans,yang2022context,DBLP:conf/emnlp/WuWZLQLH22,PPR:PPR567749}. 
To better utilize that symbolic knowledge in KGs for machine learning models, many KG embedding approaches have been devoted to representing KGs in low-dimension vector spaces \cite{DBLP:journals/tkde/WangMWG17,DBLP:conf/iclr/ZhangBL0HDZLC22}.
Traditional KG embedding models, e.g., TransE \cite{TransE}, RotatE \cite{RotatE}, are naturally taxonomized as structure-based methods \cite{DBLP:conf/aaai/XieLJLS16,DBLP:conf/aaai/ZhangCZW20,DBLP:conf/www/ZhangDSCZC20,LMKE}. These approaches employ supervised machine learning to optimize target objectives by utilizing scoring functions that preserve the inherent structure of KGs.

However, a recent shift in KG embedding methodologies has emerged, moving away from the explicit modeling of structure \cite{KG-Bert,DBLP:conf/emnlp/ZhangLZ00H20,LMKE,SimKGC}. 
Instead, contemporary techniques focus on incorporating text descriptions through the use of expressive black-box models, e.g., pre-trained language models. This new paradigm operates under the assumption that the model will inherently capture the underlying structure without requiring explicit instruction.
Leveraging language models to frame KG embeddings has emerged as a highly promising approach, yielding considerable empirical success. This technique offers the potential to generate informative representations for long-tail entities and those on-the-fly emerging entities.
However, KG embeddings with language models are usually deployed as static artifacts, which are challenging to modify without re-training.
To respond to changes (e.g., emerging new facts) or a correction for facts of existing KGs, the ability to conduct flexible knowledge updates to KG embeddings \textbf{after deployment} is desirable.

To address this need, we introduce a new task of editing language model-based KG embeddings, which aims to enable data-efficient and fast updates to KG embeddings for a small region of parametric space without influencing the performance of the rest.
It's important to note that while editing KG embeddings primarily centers on link prediction tasks, the editing of standard language models mainly addresses tasks such as question answering, each presenting its unique set of challenges.
For instance, KG embeddings contend with the complex issue of handling numerous many-to-many (N-M) triples.
Intuitively, we define two tasks for pre-trained KG embeddings, namely, EDIT and ADD, to support editing incorrect or adding new facts without re-training the whole model during deployment as shown in Figure \ref{fig:intro}.
We treat editing the memories of KG embeddings as a \emph{learning-to-update} problem, which can ensure that knowledge represented in KG embeddings remains accurate and up-to-date.
To evaluate the performance of the proposed task, we establish three guiding principles:
\textbf{Knowledge Reliability}, which indicates that those edited or newly added knowledge should correctly be inferred through link prediction; 
\textbf{Knowledge Locality}, which means that editing KG embeddings will not affect the rest of the acquired knowledge when successfully updating specific facts;
\textbf{Knowledge Efficiency} indicates being able to modify a model with low training resources.
Specifically, we build four new datasets based on FB15k237 and WN18RR for EDIT and ADD tasks. 
We leverage several approaches, including  Knowledge Editor (KE) \cite{KnowledgeEditor}, MEND \cite{MEND}, and CALINET \cite{CALINET} as baselines.
Editing KGE proves more challenging than refining language models. Experimental results demonstrate that current methods struggle with efficiently modifying KG embeddings.
We further propose a simple yet effective strong baseline dubbed \textbf{K}nowledge \textbf{G}raph \textbf{E}mbeddings E\textbf{ditor} (\ours), which can efficiently manipulate knowledge in embeddings by editing additional parametric layers.
Our experiments demonstrate that {\ours} can effectively modify incorrect knowledge or add new knowledge while preserving the integrity of other information.
We summarize our contributions as follows:
\begin{itemize}
    \item We propose a new task of editing language model-based KG embeddings.
    The proposed task with datasets may open new avenues for improving KG embedding via knowledge editing.
    
    \item We introduce \ours~that can efficiently modify incorrect knowledge or add new knowledge without affecting the rest of the acquired knowledge.
    
    \item  We conduct extensive comparisons with in-depth analysis of four datasets and report empirical results with insightful findings, demonstrating the effectiveness of the proposed approach.
\end{itemize}

\section{Related Work}
\subsection{Knowledge Graph Embedding}
Early KG embedding approaches primarily focused on deriving embeddings from structured information alone.
Most existing KG embedding methods utilize the translation-based paradigm such as TransE \cite{TransE}, TransR \cite{TransR}, TransH \cite{TransH}, and RotatE~\cite{RotatE} or semantic matching paradigms, including DistMult \cite{DistMult}, RESCAL \cite{RESCAL}, and HolE \cite{HoIE}.
Additionally, there has been significant interest in explicitly utilizing structural information through graph convolution networks \cite{GCN,GAT,CompGCN,DBLP:conf/nips/LiuGHK21,DBLP:conf/www/Zhang0YW22}.
Since pre-trained language models (PLMs) \cite{DBLP:journals/corr/abs-2303-18223} have made waves for widespread tasks, framing KG embedding via language models is an increasing technique that has led to empirical success \cite{DBLP:journals/corr/abs-2306-08302}.
On the one hand, several works have emerged to leverage PLMs for KG embeddings.
Some studies \cite{KG-Bert,DBLP:conf/emnlp/ZhangLZ00H20,Kepler,SimKGC} leverage finetuning the PLMs, for example, KG-BERT \cite{KG-Bert}, which takes the first step to utilize BERT \cite{Bert} for KG embeddings by regarding a triple as a sequence and turning link prediction into a sequence classification task.
StAR \cite{DBLP:conf/www/WangSLZW021} proposes a structure-augmented text representation approach that employs both spatial measurement and deterministic classifier for KG embeddings, respectively.
LMKE \cite{LMKE} adopts language models which utilize description-based KG embedding learning with a contrastive learning framework.
On the other hand, some studies \cite{lv-etal-2022-pre} adopt the prompt tuning \cite{liu-etal-2022-p,DBLP:conf/www/ChenZXDYTHSC22,DBLP:conf/iclr/ZhangLCDBTHC22} with language models. 
PKGC \cite{lv-etal-2022-pre}  converts triples into natural prompt sentences for KG embedding learning. 
There are also some other studies \cite{DBLP:conf/www/XieZLDCXCC22,DBLP:conf/acl/SaxenaKG22,DBLP:conf/coling/ChenWLL22} formulate KG embedding learning as sequence-to-sequence generation with language models.
Nonetheless, KG embeddings in previous studies are usually deployed as static artifacts whose behavior is challenging to modify after deployment.
To this end, we propose a new task of editing language model-based KG embeddings regarding the correction or changes for facts of existing KGs, which is the first work focusing on this to the best of our knowledge.

\subsection{Editing Factual Knowledge}
Editable training \cite{Editable_Neural_Network, yao2023editing} represents an early, model-agnostic attempt to facilitate rapid editing of trained models.
Subsequently, numerous reliable and effective approaches have been proposed to enable model editing without the necessity for resource-intensive re-training \cite{DBLP:journals/corr/abs-2307-12976, 
 DBLP:journals/corr/abs-2304-14767, DBLP:journals/corr/abs-2301-04213,HAN2023110826}.
These methods can be broadly classified into two distinct categories: external model-based editor and additional parameter-based editor.
The first category, external model-based editors, employs extrinsic model editors to manipulate model parameters.
For instance, KnowledgeEditor (KE) \cite{KnowledgeEditor} adopts a hyper-network-based approach to edit knowledge within language models. 
MEND \cite{MEND} utilizes a hypernetwork (MLP) to predict the rank-1 decomposition of fine-tuning gradients. 
The predicted gradients are used to update a subset of the parameters of PLMs.
The second category, additional parameter-based editors, involves using supplementary parameters to adjust the final output of the model, thereby achieving model editing.
Building upon prior research \cite{KnowledgeNeurons}, which suggests that Feed-Forward Networks (FFNs) in PLMs may store factual knowledge, CALINET \cite{CALINET} enhances a specific FFN within the PLM by incorporating additional parameters composed of multiple calibration memory slots. Existing approaches predominantly focus on modifying knowledge within pre-trained language models, constraining the manipulation and interpretation of facts.
In contrast, by editing knowledge in KG embeddings, our proposed tasks enable adding and modifying knowledge, thereby extending the applicability of KG embeddings across various downstream tasks.
Previous methods for updating knowledge graph embeddings (KGE), such as OUKE \cite{OUKE} and RotatH \cite{RotatH}, mainly focused on modifying the entity and relation vectors in score function-based KGE models. 
These methods employed different dataset versions as snapshots to represent changes within the knowledge graph and validated these changes directly using hit@k. However, this approach falls short of effectively measuring the efficacy of edits.
Our paper improves these methods by creating datasets and introducing metrics for a more accurate evaluation of KGE updates.

\section{Methodology}
\subsection{Task Definition}
A KG can be represented as $\mathcal{G} = (\mathcal{E}, \mathcal{R}, \mathcal{T})$, where $\mathcal{E}$, $\mathcal{R}$ and $\mathcal{T}$ are sets of entities, relation types, and triples.
Each triple in $\mathcal{T}$ takes the form $(h, r, t)$, where $h, t \in \mathcal{E}$ are the head and tail entities.
For the EDIT task, knowledge requiring edits is defined as $(h, r, y, a)$ or $(y, r, t, a)$, where $y$ is an incorrect/outdated entity.
We denote the original input entity as $x$ and the predicted target as $y$.
$a$ points to the desired edited entity (change from $y$ to $a$).
E.g., \texttt{<Donald Trump, president\_of, U.S.A.>} is an outdated triple, which should be changed to \texttt{<Joe Biden, president\_of, U.S.A.>} to update the KG embeddings.
Apart from outdated information, numerous new facts may be absent in current KGs. 
Consequently, it becomes essential to flexibly incorporate these new triples into the KG embeddings.
Thus, we introduce the ADD task to integrate new knowledge seamlessly.
Note that the ADD task is similar to the inductive setting in KG completion but without re-training the model.
Formally, we define the task of editing language model-based KG embeddings as follows:
\begin{equation}
p(a \mid \mathcal{\tilde W}, \phi, x) \gets p(y \mid \mathcal{W}, \phi, x)
\end{equation}
where $\mathcal{W}$ and $\mathcal{\tilde W}$ denote the original and edited parameters of KG embeddings, respectively.
In this paper, $x$ refers to the original input (e.g., $(?, h, r)$), output entity ($y$), and desired edited entity ($a$). 
$\phi$ represents the external parameters of the editor network.
At the same time, we need to strive to maintain the stability of the model for other correct knowledge, as described below:
\begin{equation}
p(y' \mid \mathcal{\tilde W}, \phi, x') \gets p(y' \mid \mathcal{W}, \phi, x')
\end{equation}
where $x'$ and $y'$ represent the input and label of the factual knowledge stored in the model, respectively, with $x' \neq x$.

\paragraph{Evaluation Metrics}
We set three principles to measure our proposed tasks' efficacy.

\textbf{Knowledge Reliability} evaluates whether edited or new knowledge is correctly inferred through link prediction.
Changes in KG embeddings should align with the intended edits without bringing unintended biases/errors. 
For editing effect validation, we adopt the KG completion setting. By ranking candidate entity scores, we generate an entity list. We define the \textbf{Success@1 metric (Succ@k)} by counting the number of correct triples that appear at position $k$.

\textbf{Knowledge Locality} seeks to evaluate whether editing KG embeddings will impact the rest of the acquired knowledge $\mathcal{O}^x$ when successfully updating specific facts. 
We introduce the \textbf{Retain Knowledge (RK@k)} metric, indicating that the entities predicted by the original model are \emph{still correctly inferred by the edited model}. 
Specifically, we sample the rank $k$ triples predicted by the original model as a reference dataset (L-test in Table \ref{dataset}). 
We posit that if the predicted triples remain unchanged after editing, then the model adheres to knowledge locality and does not affect the rest of the facts.
We calculate the proportion of retaining knowledge as a measure of the stability of the model editing:
\begin{equation}
     RK@k = \frac{\sum f(x';\mathcal{\tilde W})_{\le k}}{\sum f(x';\mathcal{W})_{\le k}} 
\label{RK}
\end{equation}
where $x'$ are randomly sampled from the reference dataset.
$f$ is the function with input $x$ and parameterized by $\mathcal{W}$, computing the rank of $x$. $f(x';\mathcal{W})_{\le k}$ signifies the count of rank value under $k$ predicted by the model with $\mathcal{W}$ parameters.

Besides, to better present the effect of models, we introduce two additional metrics. 
\textbf{Edited Knowledge Rate of Change} $ER_{roc}$ and \textbf{Retaining Knowledge Rate of Change} $RK_{roc}$ are denoted as:
\begin{equation}
\begin{aligned}
ER_{roc} = \frac{\left | R_{edit} - R_{origin} \right |}{R_{origin}},  \\
RK_{roc} = \frac{\left | R_{s\_edit} - R_{s\_origin} \right |}{R_{s\_edit}} 
\end{aligned}
\end{equation}
where $R_{edit}$ and $R_{origin}$ represent the mean rank on the test set before and after editing. 
$R_{s\_edit}$ and $R_{s\_origin}$ denote the reference L-test's mean rank before and after editing.

\textbf{Knowledge Efficiency} aims to evaluate the efficiency of the editors. 
We leverage the tuning parameters (including external model or additional parameters; see Params in Table \ref{edit}, \ref{add}) as a metric for evaluation.

\subsection{Datasets Construction}
\begin{figure}[htbp]
    \centering
    \includegraphics[width=0.47\textwidth]{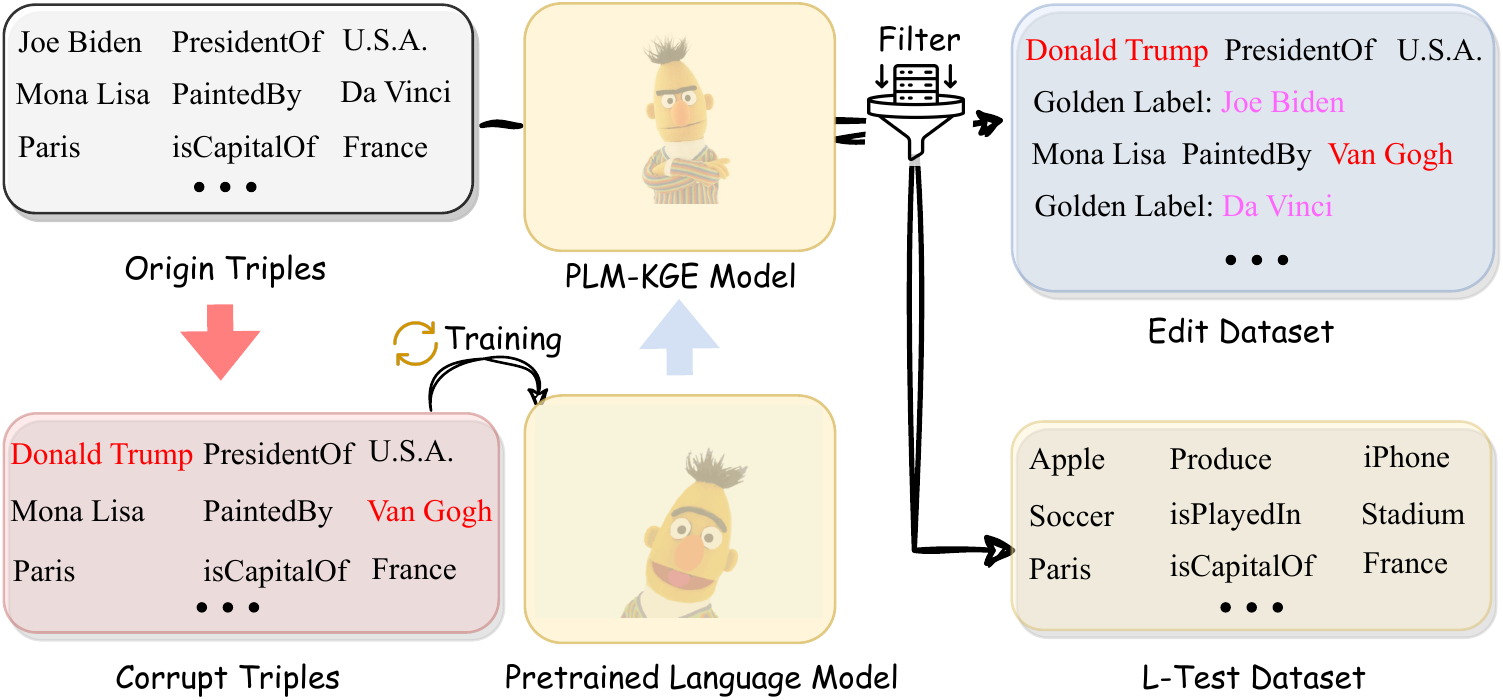}
    \caption{
    Data Construction Process. \textbf{Step 1:} Randomly disrupt the triples from the existing knowledge graph to produce a corrupted dataset. \textbf{Step 2:} Employ this dataset to fine-tune a pre-existing pre-trained KGE model, yielding a model needing editing. \textbf{Step 3:} Filtering, 
    we reassess the data with the pre-trained KGE model, accurately sorting the correctly labeled data into the L-Test dataset and allocating the mislabeled data to a designated dataset for correction.
    }
    \label{fig:dataset}
\end{figure}
We construct four datasets for EDIT and ADD tasks, based on two benchmarks FB15k-237 \cite{FB15k-237}, and WN18RR \cite{WN18RR}.
After initially training KG embeddings with language models, we sample challenging triples as candidates, as detailed below.
Recent research indicates that pre-trained language models can learn significant factual knowledge \cite{DBLP:conf/emnlp/PetroniRRLBWM19,DBLP:conf/acl/CaoLHSYLXX20}, and predict the correct tail entity with a strong bias. This may hinder the proper evaluation of the editing task. Thus, we exclude relatively simple triples inferred from language models' \textbf{internal knowledge} for precise assessment.
Specifically, we select data with link prediction ranks \textgreater 2,500.
For EDIT, the Top-1 facts from the origin model replace the entities in the training set of FB15k-237 to build the pre-train dataset (original pre-train data), while golden facts serve as target edited data. 
For ADD, we leverage the original training set of FB15k-237 to build the pre-train dataset (original pre-train data) and the data from the inductive setting as they are not seen before.
Unlike EDIT, for ADD, since new knowledge is not present during training, we \textbf{directly use the training set to evaluate the ability to incorporate new facts}. 
We adopt the same strategy to construct two datasets from WN18RR. 
To assess knowledge locality, we further create a reference dataset (L-Test) based on the link prediction performance with a rank value below $k$ (the same $k$ as in the RK metric).
Five human experts \textbf{scrutinize all datasets} for ethical concerns.
Ultimately, we establish four datasets: E-FB15k237, A-FB15k237, E-WN18RR, and A-WN18RR. 
Table \ref{dataset} provides details of the datasets. 
The creation of the datasets for the EDIT task is shown in Figure \ref{fig:dataset}.

\subsection{Language model-based KGE} \label{KGEs}
\begin{figure*}[!htbp] 
\centering 
\includegraphics[width=0.98\textwidth]{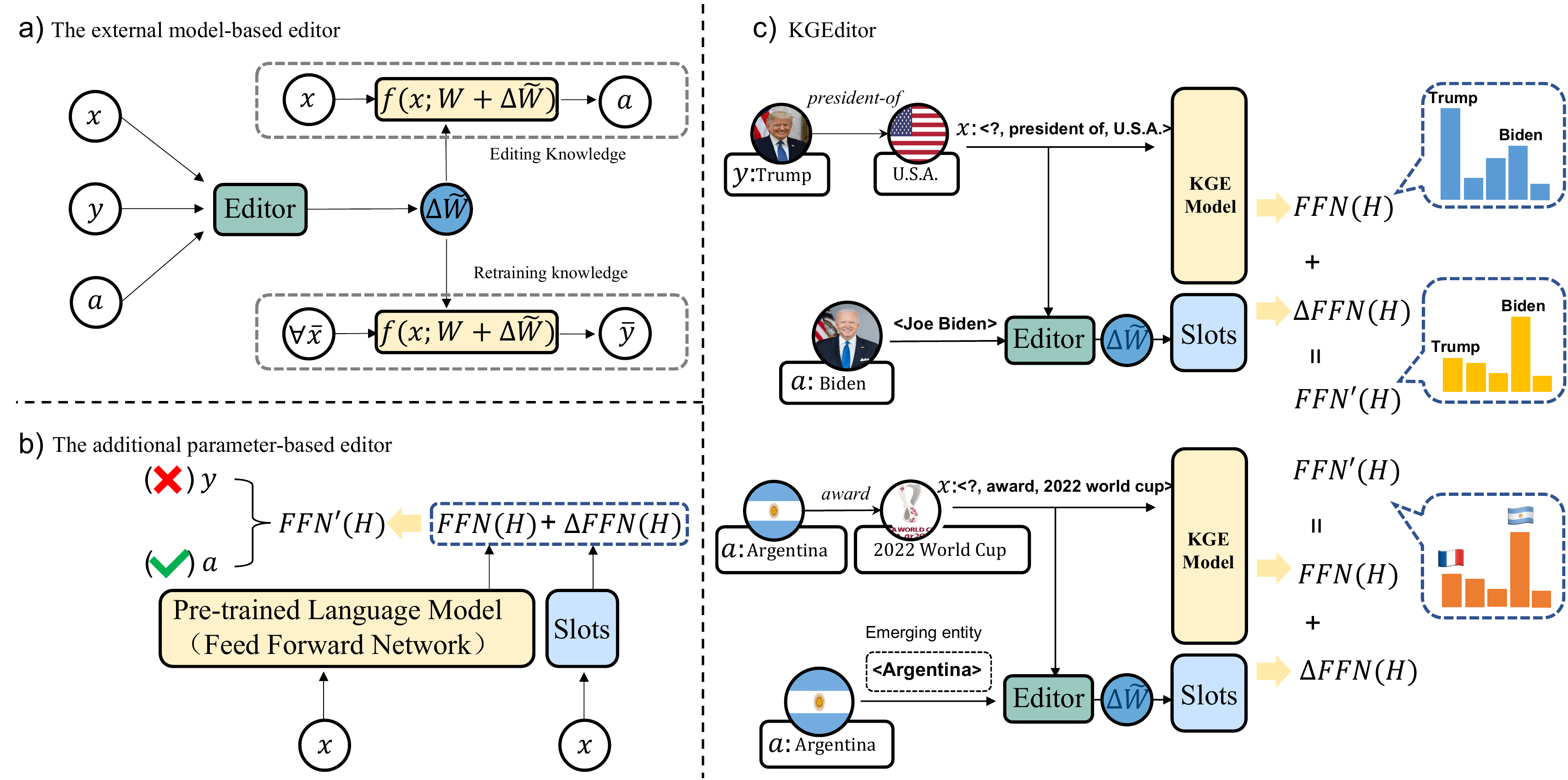} 
\caption{
The introduction of baselines and {\ours}. 
The external model-based editors 
(a) utilize a hyper external network to obtain the parameters’ shift and add to the original model parameters for editing (replacing the origin entity $y$ with the alternative entity $a$). 
The additional parameter-based editors 
(b) rectify the erroneous knowledge stored in FFN by adjusting its predicted distributions from $\operatorname{FFN}(H;W)$ to $\operatorname{FFN}^{\prime}(H)$.
{\ours} (c) utilize a hyper external network to update the knowledge in FFN (Top right: EDIT, Bottom right: ADD).}
\label{fig:model}
\end{figure*}

This section introduces the technical background of language model-based KG embeddings. 
Specifically, we illustrate two kinds of methodologies to leverage language models, namely: finetuning (\textbf{FT-KGE}) and prompt tuning (\textbf{PT-KGE}).
Note that our task is orthogonal to language-based KG embeddings, performance differences between \textbf{FT-KGE} and \textbf{PT-KGE} are discussed in the next section.

\textbf{FT-KGE} methods, such as KG-BERT \cite{KG-Bert}, which regard triples in KGs as textual sequences. 
Specifically, they are trained with the description of triples that represent relations and entities connected by the [SEP] and [CLS] tokens and then take the description sequences as the input for finetuning.
Normally, they use the presentation of [CLS] token to conduct a binary classification, note as:
\begin{equation}
    p(\mathbf{y}_{\tau} | x) = p(\texttt{[CLS]} | x ; \mathcal{W})
\label{eq:p_mem_ft}
\end{equation}
where $\mathbf{y}_{\tau} \in [0,1]$ is the label of whether the triple is positive.

\begin{table}[t]
\centering
\setlength{\tabcolsep}{3.1mm}{
\resizebox{0.5\textwidth}{!}{
\begin{tabular}{lcccccccc}
\toprule
              &         \multicolumn{4}{c}{\bf FB15k-237}                           &        \multicolumn{4}{c}{\bf WN18RR}                           \\ \cmidrule(lr){2-5}                         \cmidrule(lr){6-9}   
TASK  & Pre-train & Train & Test & L-Test& Pre-train & Train & Test & L-Test \\
\midrule
ADD     & 215,082 & 2,000 & - & 16,872 & 69,721 & 2,000 & - & 10,000 \\
EDIT    &  310,117 & 3,087 & 3,087 & 7,051 & 93,003 & 1,491 & 830 & 5,003 \\
\bottomrule
\end{tabular}
}
}

\caption{
The statistic of datasets for the EDIT and ADD tasks.
L-Test is the test set for knowledge locality to evaluate the rest of the knowledge in KGE.
}
\label{dataset}%
\end{table}

\textbf{PT-KGE} methods like PKGC \cite{lv-etal-2022-pre}, utilize natural language prompts to elicit knowledge from pre-trained models for KG embeddings. 
Prompt tuning uses PLMs as direct predictors to complete a cloze task, linking pre-training and finetuning. 
In this paper, we implement PT-KGE with entity vocabulary expansion, treating each unique entity as common knowledge embeddings. 
Specifically, we consider entities $e \in \mathcal{E}$ as special tokens in the language model, turning link prediction into a masked entity prediction.
Formally, we can obtain the correct entity by ranking the probability of each entity in the knowledge graph as follows:
\begin{equation}
    p(\mathbf{y} | x) = p(\texttt{[MASK]} = y | x ; \mathcal{W})
\label{eq:p_mem_pt}
\end{equation}

where $p(\mathbf{y} | x)$ is denoted as the probability distribution of the entity in the KG.

\subsection{Editing KGE baselines}

With the pre-trained KG embeddings via the sections mentioned above, we introduce several baselines to edit KG embeddings. 
Concretely, we divide baseline models into two categories. 
The first is the \textbf{external model-based editor}, which uses an extrinsic model editor to control the model parameters such as KE and MEND. 
Another is the \textbf{additional parameter-based editor}, which introduces additional parameters to adjust the final output of the model to achieve the model editing. 
There are also some non-editing methods in the baseline: Finetune and K-Adapter~\cite{K-adapter}.

\subsubsection{External Model-based Editor}
This method employs a hypernetwork to learn the weight update $\Delta$ for editing the language model.
KE utilizes a hypernetwork (specifically, a bidirectional-LSTM) with constrained optimization, which is used to predict the weight update during inference. 
Thus, KE can modify facts without affecting the other knowledge in parametric space.
MEND conducts efficient local edits to language models with a single input-output pair, which learns to transform the gradient of finetuned language models, which utilizes a low-rank decomposition of gradients. Hence, MEND can edit the parameters of large models, such as BART \cite{BART}, GPT-3 \cite{GPT-3}, and T5 \cite{T5}, with little resource cost.
To apply KE and MEND to our task setting, we remove the semantically equivalent set\footnote{Semantic equivalence refers to a declaration that two data elements from different vocabularies contain data having a similar meaning \cite{KnowledgeEditor}.} of the original task and only utilize the triples that need to be edited as input $x$. 

\subsubsection{Additional Parameter-based Editor}

Since previous study \cite{KnowledgeNeurons} illustrates that the FFNs in PLMs may store factual knowledge, it is intuitive to edit models by modifying the FFNs.
This paradigm introduces extra trainable parameters within the language models.
CALINET extends the FFN with additional knowledge editing parameters, consisting of several calibration memory slots.
To apply CALINET to the proposed task, we leverage the same architecture
with FFN but with a smaller intermediate dimension $d$ and add its output to the original FFN output as an adjustment term to edit knowledge.

\subsection{The Proposed Strong Baseline: {\ours}}

\begin{table*}[t]
\centering
\setlength{\tabcolsep}{3.1mm}{
\resizebox{0.98\textwidth}{!}{
\begin{tabular}{lccccccccccccc}
\toprule
\multirow{2}{*}{Method}    &    \multirow{2}{*}{Params}    &         \multicolumn{6}{c}{\bf E-FB15k237}                           &        \multicolumn{6}{c}{\bf E-WN18RR}                           \\ \cmidrule(lr){3-8}   \cmidrule(lr){9-14}   
   &     & Time $\downarrow$ &  Succ@1 $\uparrow$     &     Succ@3 $\uparrow$      &  $ER_{roc}$ $\uparrow$ &    RK@3 $\uparrow$ & $RK_{roc}$ $\downarrow$ & Time $\downarrow$ &   Succ@1 $\uparrow$ &  Succ@3 $\uparrow$      &  $ER_{roc}$ $\uparrow$ & RK@3 $\uparrow$ & $RK_{roc}$ $\downarrow$ \\ 
\midrule      
\multicolumn{14}{c}{\textit{No Model Edit}}                                                                              \\
\midrule
KGE\_FT     &  121M & 0.103 & 0.472 & 0.746 & 0.998 & 0.543 & 0.977 & 0.109 & 0.758 & 0.863 & 0.998 & 0.847 & 0.746 \\
KGE\_ZSL    & 0M & 0.000 & 0.000 & 0.000 & - & 1.000 & 0.000 & 0.000 & 0.000	&  0.000 & - & 1.000 & 0.000 \\
K-Adapter & 32.5M & 0.056 & 0.329 & 0.348 & 0.926 & 0.001 & 0.999 & 0.061 & 0.638 & 0.752 & 0.992 & 0.009 & 0.999 \\
\midrule
\multicolumn{14}{c}{\textit{Model Edit Method}}                                                                   \\
\midrule
CALINET   & 0.9M & \underline{0.257} &	0.328 & 0.348 & 0.937 & 0.353 & 0.997 & \underline{0.238} & 0.538 & 0.649 & \textbf{0.991} & 0.446 & 0.994 \\
KE  & 88.9M & 0.368 & 0.702 & \underline{0.969} & \textbf{0.999} & \textbf{0.912} & \underline{0.685} & 0.386 &  0.599 & 0.682 & \underline{0.978} & 0.935 & \textbf{0.041} \\
MEND &  59.1M & 0.280 & \underline{0.828} & 0.950 & \underline{0.954} & 0.750 & 0.993 & 0.260 & \underline{0.815} & \underline{0.827} & 0.948 & \textbf{0.957} & 0.772  \\
\midrule
\ours    & 38.9M  & \textbf{0.226} & \textbf{0.866} & \textbf{0.986} & \textbf{0.999} & \underline{0.874} & \textbf{0.635} & \textbf{0.232} & \textbf{0.833}	& \textbf{0.844}	& \textbf{0.991} & \underline{0.956} & \underline{0.256} \\
\bottomrule
\end{tabular}
}
}
\caption{The main result of the EDIT task on E-FB15k237 and E-WN18RR. \textbf{Bold} indicates the best result among \textit{Model Edit Method}, while \underline{underline} represents the second-best result. The same format applies to Table \ref{add}.}
\label{edit}
\end{table*}

\begin{table*}[t]
\centering
\setlength{\tabcolsep}{3.1mm}{
\resizebox{0.98\textwidth}{!}{
\begin{tabular}{lccccccccccccc}
\toprule
\multirow{2}{*}{Method}    &    \multirow{2}{*}{Params}    &         \multicolumn{6}{c}{\bf A-FB15k237}                           &        \multicolumn{6}{c}{\bf A-WN18RR}                           \\ \cmidrule(lr){3-8}   \cmidrule(lr){9-14}    
   &     & Time $\downarrow$ & Succ@1 $\uparrow$     &     Succ@3 $\uparrow$      &  $ER_{roc}$ $\uparrow$ &    RK@3 $\uparrow$ & $RK_{roc}$ $\downarrow$  & Time $\downarrow$ &  Succ@1 $\uparrow$ &  Succ@3 $\uparrow$      &  $ER_{roc}$ $\uparrow$ & RK@3 $\uparrow$ & $RK_{roc}$ $\downarrow$ \\ 
\midrule      
\multicolumn{14}{c}{\textit{No Model Edit}}                                                                              \\
\midrule

KGE\_FT     &  121M & 0.100 & 0.906 & 0.976 & 0.999 & 0.223 & 0.997 & 0.108 & 0.997 & 0.999 & 0.999 &  0.554 & 0.996 \\
KGE\_ZSL    & 0M & 0.000 & 0.000 & 0.000 & - & 1.000 & 0.000 & 0.000 & 0.000 &  0.000 & - & 1.000 & 0.000 \\
K-Adapter    & 32.5M & 0.055 & 0.871 & 0.981 & 0.999 & 0.000 & 0.999 & 0.061 & 0.898 & 0.978 & 0.999 & 0.002 & 0.999 \\
\midrule
\multicolumn{14}{c}{\textit{Model Edit Method}} \\
\midrule
CALINET    & 0.9M & \underline{0.261} &	\underline{0.714} & 0.870 & \underline{0.997} & 0.034 & 0.999 & \underline{0.275} & 0.832 & 0.913 & \underline{0.995} & 0.511 & 0.989 \\
KE & 88.9M & 0.362 & 0.648 & \underline{0.884} & \underline{0.997} & \textbf{0.926} & \underline{0.971} & 0.384 & 0.986 & \underline{0.996} & \textbf{0.999} & \textbf{0.975} & \textbf{0.090} \\
MEND    & 59.1M & 0.400 & 0.517 & 0.745 & 0.991 & 0.499 & 0.977 & 0.350 & \textbf{0.999} & \textbf{1.0} & \textbf{0.999} & 0.810  & 0.987 \\
\midrule
\ours    & 58.7M & \textbf{0.203} & \textbf{0.796} & \textbf{0.923} & \textbf{0.998} & \underline{0.899} & \textbf{0.920} & \textbf{0.203} & \underline{0.998} & \textbf{1.0} & \textbf{0.999} & \underline{0.956} & \underline{0.300} \\
\bottomrule
\end{tabular}
}
}
\caption{The main result of the ADD task on A-FB15k237 and A-WN18RR.}
\label{add}
\end{table*}

Note that external model-based edit methods are flexible but have to optimize many editor parameters.
In contrast, the additional parameter-based method only tunes a small number of additional parameters (0.9M) but encounters poor empirical performance due to the challenge of manipulating knowledge in PLMs.
Intuitively, we capitalize on the advantages of both approaches and propose a simple yet robust baseline {\ours}, as shown in Figure \ref{fig:model}, which employs additional parameters through hypernetwork.
We construct an additional layer with the same architecture of FFN and leverage its parameters for knowledge editing.
Moreover, we leverage an additional hypernetwork to generate the additional layer.
With {\ours}, we can optimize fewer parameters while keeping the performance of editing KG embeddings.
Specifically, when computing the output of FFN, we add the output of the additional FFNs to the original FFN output as an adjustment term for editing:
\begin{equation}
\small
    FFN'(H) = FFN(H;\mathcal{W}) + \bigtriangleup FFN(H;\mathcal{\tilde W} + \bigtriangleup \mathcal{\tilde W})
\label{FFN}
\end{equation}

where $\mathcal{W}$ is denoted as the origin FFNs' weight, and $\mathcal{\tilde W}$ is the weight of additional FFNs.
$\bigtriangleup \mathcal{\tilde W}$ is the parameters’ shift generated by the hypernetwork. 
Inspired by  KE \cite{KnowledgeEditor}, we build the hypernetworks with a bidirectional LSTM.
We encode  $\langle x, y, a \rangle$ and then concatenate them with a special separator to feed the bidirectional LSTM.
Then, we feed the last hidden states of bidirectional LSTM into the FNN to generate a single vector $h$ for knowledge editing.
To predict the shift for the weight matrix $\mathcal{\tilde W}^{n \times m}$, we leverage FNN conditioned on $h$ which predict vectors $\alpha, \beta \in \mathbb{R}^m, \gamma, \delta \in \mathbb{R}^n$ and a scalar $\eta \in \mathbb{R}$. 
Formally, we have the following:
\begin{equation}
\small
\begin{aligned}
    \bigtriangleup \mathcal{\tilde W} = \sigma(\eta)\cdot \left( \hat\alpha \odot \nabla_\mathcal{W} \mathcal{L}(\mathcal{W};x,a) + \hat\beta \right) \;, \\
    \text{with} \quad \hat\alpha = \hat{\sigma}(\alpha)\;\!\gamma^\top \quad \text{and} \quad \hat\beta = \hat{\sigma}(\beta)\;\!\delta^\top \;,
\end{aligned}
\end{equation}
where $\hat{\sigma}$ refers to the Softmax function (\emph{i.e.}\xspace, $x \mapsto \exp(x) / \sum_i \exp(x_i)$) and $\sigma$ refers to the Sigmoid function (\emph{i.e.}\xspace, $x \mapsto (1 + \exp(-x))^{-1}$).
$\nabla_\mathcal{W} \mathcal{L}(\mathcal{W};x, a)$ presents the gradient, which contains rich information regarding how to access the knowledge in $\mathcal{W}$.
Note that the hypernetwork $\phi$ parameters can linearly scale with the size of $\mathcal{\tilde W}$, making the {\ours} efficient for editing KG embeddings in terms of computational resources and time.

\section{Evaluation}
\subsection{Settings}
We employ the constructed datasets, E-FB15k237, A-FB15k237, E-WN18RR, and A-WN18RR for evaluations. 
Initially, we utilize the pre-train datasets (the identical training set for the ADD task and the corrupted training set for the EDIT task) to initialize the KG embeddings with BERT. 
We then train the editor using the training dataset and evaluate using the testing dataset. 
Regarding the ADD task, the evaluation is conducted based on the performance of the training set, as they are not present in the pre-training dataset.
We adopt PT-KGE as the default setting for the main experiments.
Additionally, We supply the datasets, pre-trained KG embeddings, and a leaderboard for reference.

\subsection{Main Results}

\begin{figure}[t!]
\centering
\includegraphics[width=0.50\textwidth]{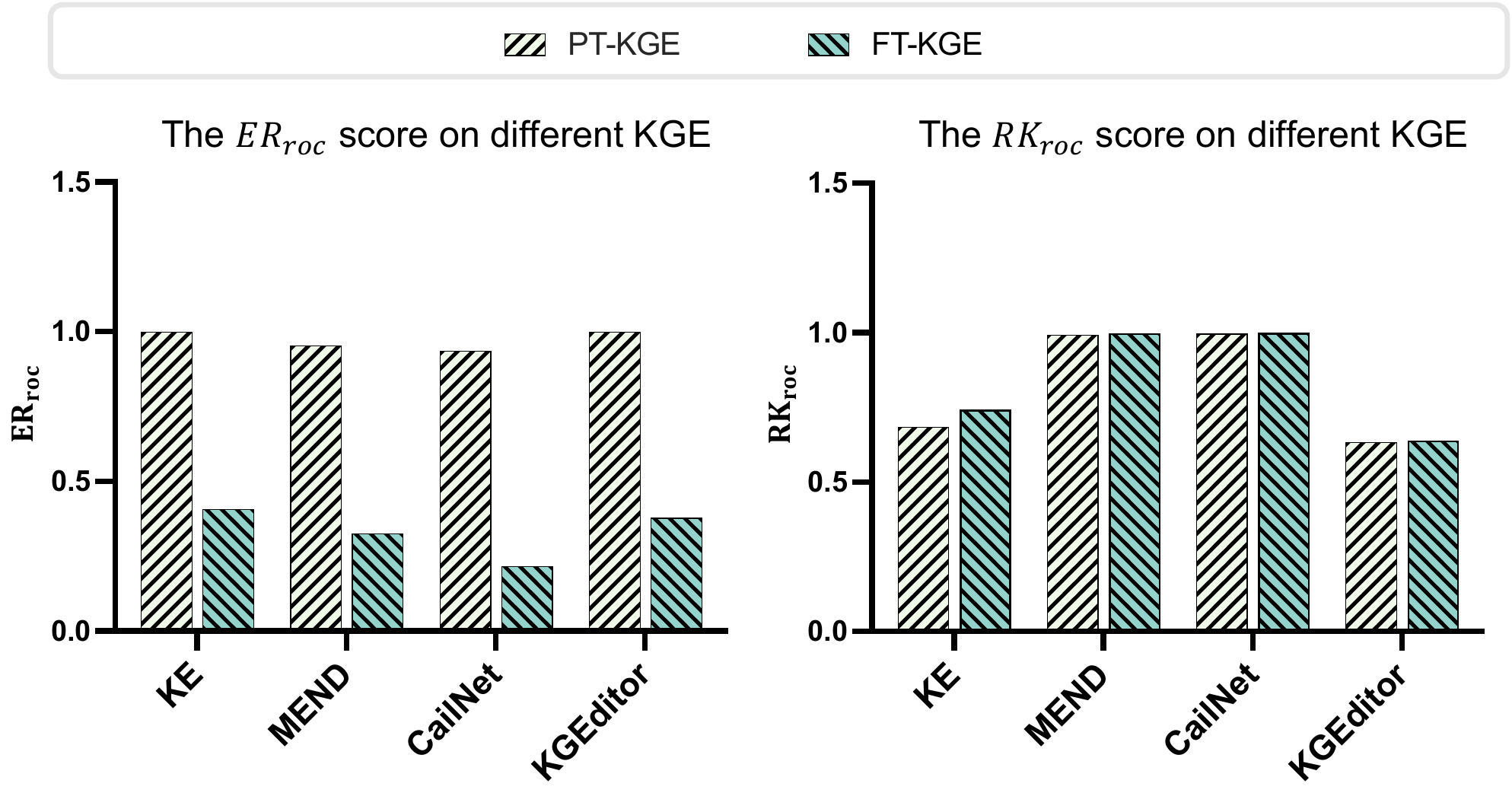}
\caption{
\textbf{Left: $ER_{roc}$} and \textbf{Right: $RK_{roc}$}. 
The performance of different KGE initialization with different models.}
\label{fig:kge}
\end{figure}

\begin{figure}[t!]
\centering
\includegraphics[width=0.50\textwidth]{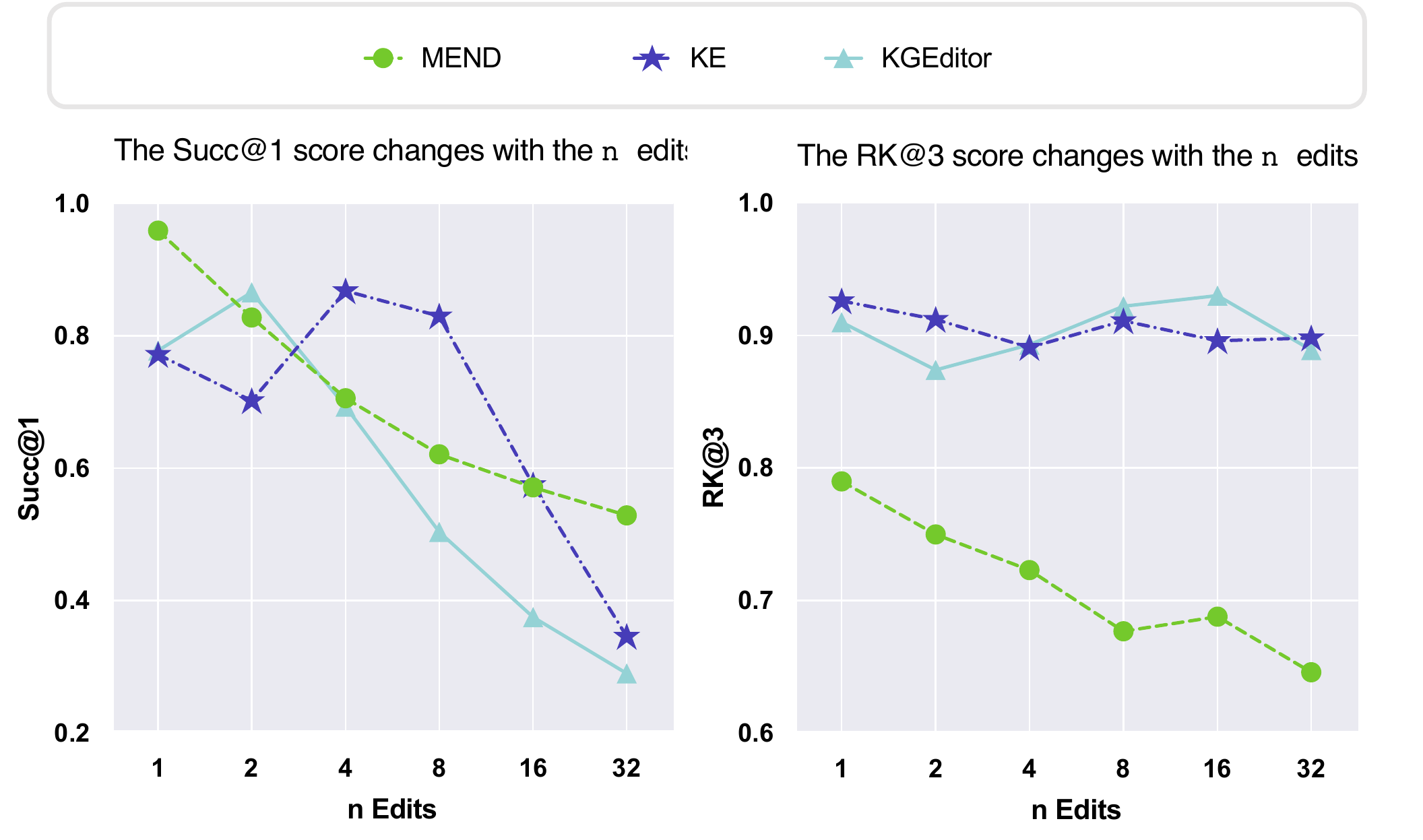}
\caption{
\textbf{Left:} the variation of Succ@1 value with $n$ edits. 
All models drop when $n$ edits become larger. 
\textbf{Right:} the RK@3 score drops when $n$ changes, KE and {\ours} can remain stable performance, while MEND decreases.}
\label{fig:num_edit}
\end{figure}

We compare {\ours} with several baselines, including MEND, KE, CALINET, as well as some variants: KGE\_FT, which involves directly fine-tuning all parameters; KGE\_ZSL, which entails directly inferring triples without adjusting any parameters; K-Adapter, which employs an adaptor to fine-tune a subset of parameters.
Our experiments aim to 1) evaluate the efficacy of various approaches for editing KG embeddings (Table \ref{edit},\ref{add}), 2) investigate the performance when varying the number ($n$) of target triples (Figure \ref{fig:num_edit}), and 3) examine the impact of distinct KGE initialization approaches on performance(Figure \ref{fig:kge}).
\subsubsection{EDIT\&ADD Task}

As Table \ref{edit} shows, KGE\_ZSL fails to infer any facts on both datasets for the EDIT task.
Intriguingly, we also find that fine-tuning all (KGE\_FT) or a subset (K-Adaptor) of parameters does not result in satisfactory performance.
We think this is due to the numerous $N$-$M$ facts, and fine-tuning with more data may lead to confusion about the knowledge. 
Nonetheless, we discover that all carefully designed knowledge editing baselines outperform KGE\_FT and K-Adaptor, suggesting that specific architectures should be developed for editing knowledge to locate better and correct facts.
Furthermore, we notice that CALIET exhibits poor knowledge locality (RK@3 and $RK_{roc}$), indicating that merely modifying a few parameters in FFNs cannot guarantee precise knowledge editing.
Our model {\ours} surpasses nearly all baselines, utilizing fewer tunable parameters than KE and MEND.
Regarding time and resource expenditure, {\ours} exhibits markedly enhanced editing efficiency compared to KGE\_FT.
Note that our model employs a hyper network to guide the parameter editing of the FFN, resulting in greater parameter efficiency than KE and MEND while simultaneously achieving superior performance to CALINET. 

For the ADD task, as shown in Table \ref{add}, we observe that KGE\_ZSL continues to fail in inferring any facts on both datasets (We filter all facts that can be directly inferred by probing the PLM).
Finetuning either all parameters (KGE\_FT) or a subset (K-Adaptor) maximizes the performance of reliability, but compromises previously acquired knowledge, reflecting poor knowledge locality.
Given that the ADD task introduces unseen facts during KG embeddings initialization, finetuning parameters naturally captures new knowledge while sacrificing prior information.

\subsubsection{FT-KGE\&PT-KGE}

We evaluate various KGE initialization methods with different models on the EDIT task. 
Figure \ref{fig:kge} contrasts the performance of FT-KGE and PT-KGE in terms of knowledge reliability and locality.
We observe that the editing performance of PT-KGE is superior to that of FT-KGE, suggesting that employing prompt-based models is more suitable for editing KG embeddings.
Prior research \cite{liu2021pre,DBLP:journals/corr/abs-2202-05262} has demonstrated that the prompt-based approaches are more effective in harnessing the ``modeledge'' \cite{DBLP:journals/aiopen/HanZDGLHQYZZHHJ21} in PLMs, making it reasonable to utilize PT-KGE for editing purposes.

\subsubsection{Number of Edited Knowledge}
We further analyze the impact of varying the number of edited facts. 
\begin{figure}[htbp]
    \centering
    \includegraphics[width=0.50\textwidth]{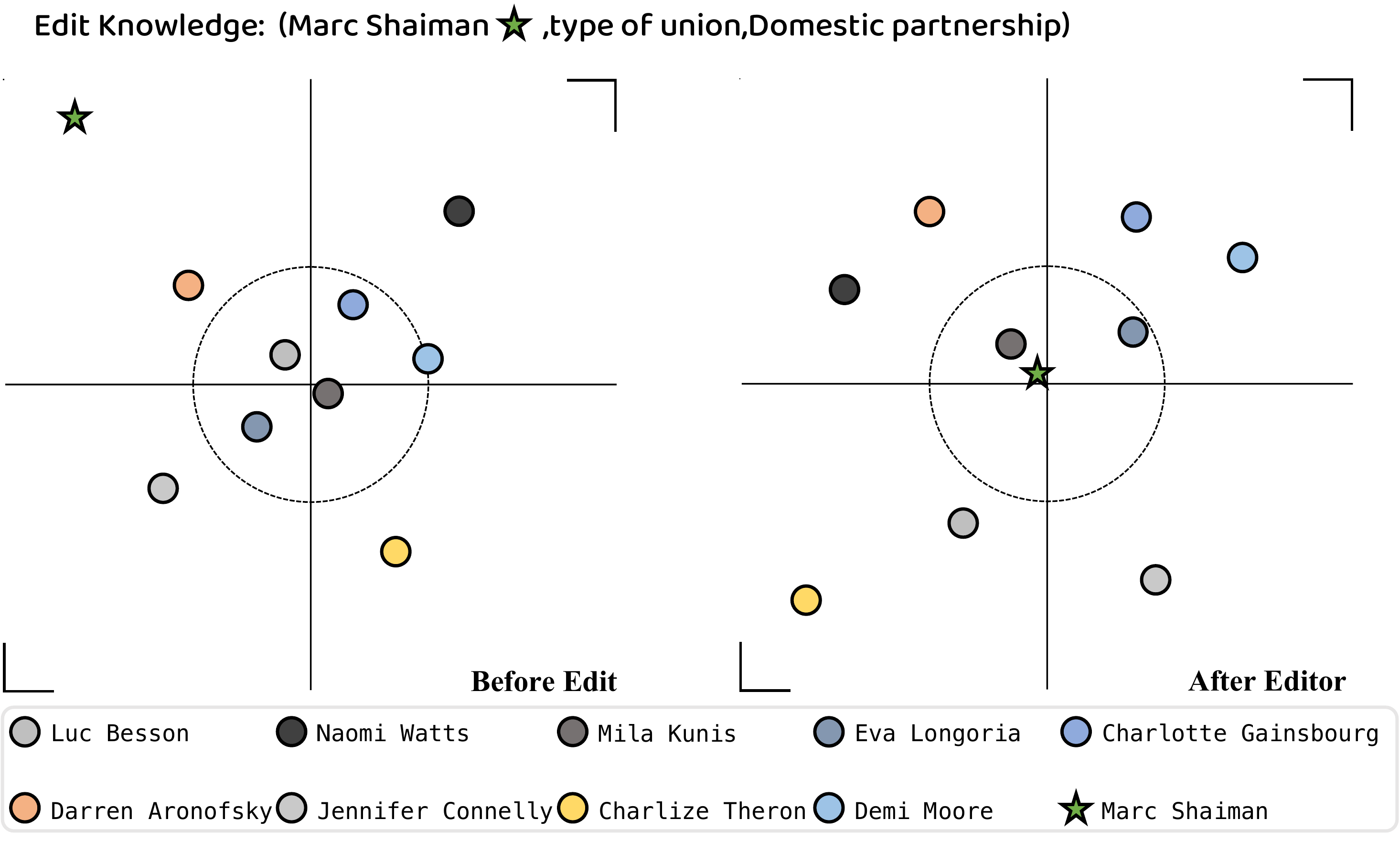}
    \caption{
    We edit the fact (\textbf{Mila Kunis}, type of union, Domestic partnership) stored in the KG embeddings and replace the head entity by \textbf{Marc Shaiman}  (Golden Label). 
    The points close to the center refer to the entities the model prefers. 
    \textbf{Left:} predictions of the original KG embeddings.
    \textbf{Right:} predictions after editing.}
    \label{fig:case}
\end{figure}
We employ different numbers of edits in the E-FB15k237, with the number of edits $n \in \{1,2,4,8,16,32\}$.
Figure \ref{fig:num_edit} illustrates a pronounced impact of $n$ edits on all models' knowledge reliability, measured by Succ@1.
Furthermore, we observe that KE and {\ours} show stable performance in knowledge locality, unlike MEND, which sees a notable decline.
We attribute this to KE and {\ours} employing FFN (termed as knowledge neurons \cite{DBLP:conf/acl/DaiDHSCW22}), which can be scaled to accommodate a higher number of edited facts.

We visualize entities before and after editing for clearer model insights. 
Figure \ref{fig:case} illustrates a substantial shift in predicted entities' positions before and after the model editor's application.
These points signify that the inferred entity is provided with a head entity and relation, with the right center representing the golden standard.
Upon editing the model, the correct entity distinctly emerges close to the circle's center (knowledge reliability), while the relevant distances of other entities remain largely unaltered, thereby showcasing the effectiveness of editing KG embeddings.

\section{Discussion and Conclusion}
Our proposed task of editing the KGE model allows for direct modification of knowledge to suit specific tasks, thereby improving the efficiency and accuracy of the editing process. 
Contrary to earlier pre-training language model tasks for editing, our approach relies on KG facts to modify knowledge, without using pre-trained model knowledge.
These methods enhance performance and offer important insights for research in knowledge representation and reasoning.

\section{Future Work}
Editing KGE models presents ongoing issues, notably handling intricate knowledge and many-to-many relations. 
Edited facts, arising from such relations, can bias the model to the edited entity, overlooking other valid entities.
Besides, the experimental KGE models are small, all leveraging the standard BERT.
Yet, with rising large-scale generative models (LLMs) like LLaMA \cite{llama}, ChatGLM \cite{ChatGLM}, and Alpaca \cite{alpaca}, the demand to edit LLMs grows.
In the future, we aim to design models to edit knowledge with many-to-many relations and integrate LLMs editing techniques.

\section{Acknowledgments}
We would like to express gratitude to the anonymous reviewers for their kind comments. This work was supported by the National Natural Science Foundation of China (No. 62206246), the Fundamental Research Funds for the Central Universities (226-2023-00138), Zhejiang Provincial Natural Science Foundation of China (No. LGG22F030011), Ningbo Natural Science Foundation (2021J190), Yongjiang Talent Introduction Programme (2021A-156-G), CCF-Tencent Rhino-Bird Open Research Fund, and Information Technology Center and State Key Lab of CAD\&CG, Zhejiang University.

\bibliography{aaai24}

\end{document}